\documentclass[11pt]{article}

\usepackage[final]{acl}

\usepackage{times}
\usepackage{latexsym}

\usepackage[T1]{fontenc}

\usepackage[utf8]{inputenc}

\usepackage{microtype}

\usepackage{inconsolata}

\usepackage{graphicx}
\usepackage{listings}
\usepackage{xcolor}
\usepackage{calc}
\usepackage{mathtools}

\usepackage{enumitem}

\lstdefinestyle{examplebox}{
  basicstyle=\ttfamily\footnotesize,
  frame=single,
  framerule=0pt,
  rulecolor=\color{black},
  backgroundcolor=\color{black!2},
  xleftmargin=1pt,
  xrightmargin=1pt,
  framexleftmargin=4pt,
  framexrightmargin=4pt,
  framextopmargin=4pt,
  framexbottommargin=4pt,
  breaklines=true,
  breakatwhitespace=true,
  columns=fullflexible,
  keepspaces=true,
  showstringspaces=false
}

\newcommand{\legalref}[1]{{\textit{#1}}\nobreak}

\newcommand{\bundesrecht}{\texttt{bundesrecht}}
\newcommand{\bundesrechtclass}{Bundesrecht}
\newcommand{\bundesrechtdataset}{Bundesrecht}

\newcommand{\boldfw}[1]{\mathmakebox[\widthof{$#1$}][c]{\pmb{#1}}}

\title{Bundesrecht: An Open Resource for\\ German Statutory Reference Processing}

\author{
  \textbf{Harshil Darji\textsuperscript{1,2}},
  \textbf{Martin Heckelmann\textsuperscript{1}},
  \textbf{Christina Kratsch\textsuperscript{1}},
  \textbf{Gerard de Melo\textsuperscript{2}}
\\
\\
  \textsuperscript{1}Hochschule für Technik und Wirtschaft Berlin, Germany,\\
  \textsuperscript{2}Hasso-Plattner Institute / University of Potsdam, Germany
\\
  \small{
    \textbf{Correspondence:} \href{mailto:Harshil.Darji@HTW-Berlin.de}{Harshil.Darji@HTW-Berlin.de}
  }
}

\begin{document}
\maketitle

\begin{abstract}
Statutory references are central to legal language understanding, but are difficult to process automatically, as they appear in compact, variable forms, may combine multiple targets, and often point to lower-level units.
Existing systems for German focus either on parsing references from legal documents or accessing statutory text once citations are explicit.
This paper introduces \bundesrecht, an open resource for German statutory reference processing, consisting of a software library and a structured corpus of German federal law.
The library parses, normalizes, and resolves German statutory references, mapping raw citation strings to structured objects, expanding compact references, and linking them to statutory provisions. The dataset preserves the internal hierarchy of statutes from laws to fine-granular subclauses.
We evaluate the parser and normalizer on $2{,}944$ annotated German legal references using strict exact-match and micro-averaged information extraction metrics, and compare against existing systems, achieving the highest per-field F1. We manually verify resolution against the official statutory text, and show that canonical reference deduplication groups citation surface variants far more reliably than string matching.
\bundesrecht{} is the first open resource covering German statutory reference processing end-to-end, from raw citation string to resolved statutory provision, and is available on PyPI.
\end{abstract}

\section{Introduction}

Legal texts rarely stand alone. Court decisions, statutes, administrative rules, and legal commentaries often refer to statutory provisions to invoke definitions, exceptions, procedures, legal consequences, or competence rules~\citep{adedjouma2014automated}. A sentence that cites \legalref{\S~823 BGB} or \legalref{Art. 3 GG} points to another statutory provision that may be necessary for understanding the text. For legal NLP, such references are therefore important for retrieval~\citep{louis2022statutory}, legal question answering and retrieval-augmented generation~\citep{louis2024interpretable,buttner2024answering}, citation extraction, corpus construction, and the creation of linked legal datasets.

German statutory references are difficult to process automatically because they combine compact legal writing with a deeply structured target system. A reference may point to a complete \textit{Gesetz} (law), a \textit{Paragraph} or \textit{Artikel}, or a more specific level such as \textit{Absatz} (paragraph), \textit{Satz} (sentence), \textit{Nummer} (number), \textit{Buchstabe} (letter), or \textit{Unterbuchstabe} (sub-letter). Their surface form also varies: references may use abbreviations such as \textit{i.V.m.}, combine several targets in one expression, inherit context across comma-separated parts, or refer to ranges using \textit{\S\S}. Proper interpretation therefore requires parsing the internal structure of a reference, expanding compact forms into individually resolvable references, and resolving each against the correct statutory text.

Existing German legal NLP resources address adjacent tasks, but statutory reference processing remains insufficiently addressed end-to-end. Legal datasets provide court decisions for downstream tasks, reference parsers extract citation strings, and other tools access statutory text only when the citation string is already precise. What is missing is an open resource that connects these steps: from a raw German statutory reference, to a canonical reference, to the resolved statutory unit in a structured corpus of German federal law.

We therefore introduce \bundesrecht, an open resource for German statutory reference processing. It consists of a Python library\footnote{\url{https://pypi.org/project/bundesrecht/}} and a corresponding structured JSONL corpus\footnote{\url{https://huggingface.co/datasets/harshildarji/bundesrecht}} of German federal laws derived from \textit{Gesetze im Internet}\footnote{\url{https://www.gesetze-im-internet.de/}}. The library parses, normalizes, and resolves German statutory references in both \textit{Paragraph}-based and \textit{Artikel}-based forms, and the corpus preserves the internal hierarchy of federal statutes down to the \textit{Unterbuchstabe} level, so that references can be resolved below the level of a \textit{Paragraph} or \textit{Artikel}.

We evaluate the parser and normalizer on $2{,}944$ manually annotated German legal references. Field-level exact match rates attain at least $99.7\%$ for all fields under strict exact-match, and micro-averaged F1 surpasses $98\%$ at all but one granularity level. In comparison with existing systems, \bundesrecht{} obtains the highest per-field F1 across all comparable fields. We also manually verify resolution against the official statutory text and find it correct for $94.5\%$ of targets in a stratified sample. We further evaluate canonical reference deduplication on observed citation surface forms from German court decisions, and show that normalized references group variants of the same statutory target far more reliably than raw strings.

\section{Related Work}
\label{sec:related}

Several resources make German legal documents easier to use in downstream tasks. \citet{ostendorff2020towards} introduced Open Legal Data, a platform for accessing German legal documents with structured metadata. \citet{wrzalik2021gerdalir} introduced \textit{GerDaLIR}, a German legal information retrieval dataset that links query passages to relevant case documents. \citet{darji2025segmentation} processed Open Legal Data into a cleaned and sectioned JSONL dataset of $251{,}038$ German court decisions, separating the text into \textit{Tenor}, \textit{Tatbestand}, and \textit{Entscheidungsgründe}. In contrast to these court-decision resources, our work focuses on German federal statutes and references to statutory provisions.

More relevant to our work, \citet{darji2021exploring} studied semantic similarity between German legal texts and the laws they refer to, showing that statutory references can connect court decisions with the provisions they cite. \citet{darji2023dataset} then introduced a dataset of German legal reference annotations, where references are manually annotated with structured fields rather than treated only as text spans. This dataset provides the structured target fields against which we evaluate \bundesrecht{}. \citet{leitner2020dataset} introduced a German legal NER dataset whose schema includes reference-related entities such as laws, ordinances, and EU legal norms. This addresses reference \textit{identification}, whereas our work targets the following step of normalizing and resolving identified references.

Currently, several public tools address German legal references or statutory text retrieval, and we compare against the three most closely related in Section~\ref{sec:tool_comparison}. The \texttt{german-legal-reference-parser}~\citep{germanlegalreferenceparser} parses references into structured objects but does not produce canonical references resolvable against statutory text. The \texttt{legal-reference-extraction} tool~\citep{legalreferenceextraction} extracts law and case references from full text. This step can precede \bundesrecht{}, which then parses, normalizes, and resolves an identified reference. The \texttt{Gesetzessuche} project~\citep{gesetzessuche} looks up statutory text from explicit citation strings such as \legalref{BGB \S~7 Absatz 1}, but does not handle compact or compound references such as \legalref{\S~312 i.V.m.\ \S~355 BGB} or \legalref{\S~2 Abs.\ 1 Nr.\ 1, Nr.\ 7, Abs.\ 2 UrhG}, which must first be expanded before they can be resolved.

Together, these efforts have made German legal texts easier to access, annotate, retrieve, and process. What remains is that many references are too compact for direct lookup: they must be converted into canonical references before they can be resolved to the correct \textit{Paragraph}, \textit{Absatz}, \textit{Nummer}, or \textit{Buchstabe}. This is the focus of \bundesrecht{}, which processes a citation string as a structured object rather than a text span or lookup string.

\section{Statutory Reference Processing}
\label{sec:reference_processing}

Statutory reference processing starts from a raw citation string and aims to produce a reference that can be used by a legal NLP system. In full legal documents, citation extraction may be needed first, and the task addressed here begins once a candidate statutory reference string has been identified. We divide this task into three steps: Parsing identifies the internal structure of a reference, normalization rewrites it into canonical and individually resolvable forms, and resolution links each reference to the corresponding statutory provision.

\subsection{Parsing}
Parsing maps a raw citation string to a structured representation. For example, \legalref{\S~2 Abs.\ 1 Nr.\ 1 UrhG} should be identified not only as a reference, but also as \textit{Paragraph} $2$ of the UrhG law, \textit{Absatz} $1$, \textit{Nummer} $1$. German references may use \textit{Paragraph}-based or \textit{Artikel}-based forms and may contain a hierarchy of internal structure, including \textit{Absatz}, \textit{Satz}, \textit{Nummer}, \textit{Buchstabe}, \textit{Alternative}, and \textit{Halbsatz}. This structure matters because two references to the same \textit{Paragraph} may point to different content if they differ at a lower level: \legalref{\S~2 Abs.\ 1 Nr.\ 1 UrhG} and \legalref{\S~2 Abs.\ 1 Nr.\ 7 UrhG} share the same \textit{Paragraph} and \textit{Absatz} but refer to different \textit{Nummer} parts, and a parser must preserve this difference.

\subsection{Normalization}
Normalization converts raw or parsed references into canonical citation strings. This is necessary because German legal references often use compact forms that are not directly resolvable: one citation string may contain several targets, omit repeated context, or use abbreviated forms that must be expanded before lookup. For example, \legalref{\S~2 Abs.\ 1 Nr.\ 1, Nr.\ 7, Abs.\ 2 UrhG} contains three targets, \legalref{\S~2 Abs.\ 1 Nr.\ 1 UrhG}, \legalref{\S~2 Abs.\ 1 Nr.\ 7 UrhG}, and \legalref{\S~2 Abs.\ 2 UrhG}, where the later parts inherit the \textit{Paragraph} and law abbreviation from the full string. Similarly, \legalref{\S~312 i.V.m.\ \S~355 BGB} connects two references via \textit{i.V.m.}, and \legalref{\S\S~12--14 BGB} refers to a range. Normalization makes these implicit targets explicit by producing a list of canonical references.

\subsection{Resolution}
Resolution maps a canonical reference to the corresponding statutory provision in a machine-readable corpus. This requires more than matching the law abbreviation and \textit{Paragraph}: if a reference contains \textit{Absatz}, \textit{Satz}, \textit{Nummer}, or \textit{Buchstabe}, the resolver must use the internal structure of the statute to identify the correct target. For example, \legalref{\S~312 Abs.\ 2 Nr.\ 1 BGB} should resolve to BGB, \textit{Paragraph} $312$, \textit{Absatz} $2$, \textit{Nummer} $1$. If the requested lower-level target is available, the resolver returns its text directly. If it cannot be matched exactly, the resolver returns the closest available provision with information about the level that was resolved, which makes resolution useful for both exact lookup and downstream tasks where partial resolution suffices.

\section{The \bundesrecht{} Library}
Our \bundesrecht{} library implements the three steps described above. It is written in Python and exposes a small set of functions and classes for parsing, normalizing, and resolving German statutory references. We have kept the three steps separate, while still allowing them to be used together in one pipeline. A user can parse a reference without loading the corpus, normalize a citation string into canonical references, or load the corpus once and resolve many references against it.

\subsection{Design Goals}
The library is built with three design goals: It should return structured objects rather than modified strings, so that downstream systems can tell whether a reference points to a \textit{Paragraph}, \textit{Absatz}, \textit{Nummer}, or \textit{Buchstabe}. It should make compact citation strings explicit by returning one canonical reference per target. And it should support corpus-based resolution, so that a canonical reference can be linked to the corresponding statutory provision.

\subsection{Architecture}

\begin{figure}[h!]
    \centering
    \includegraphics[width=0.75\columnwidth]{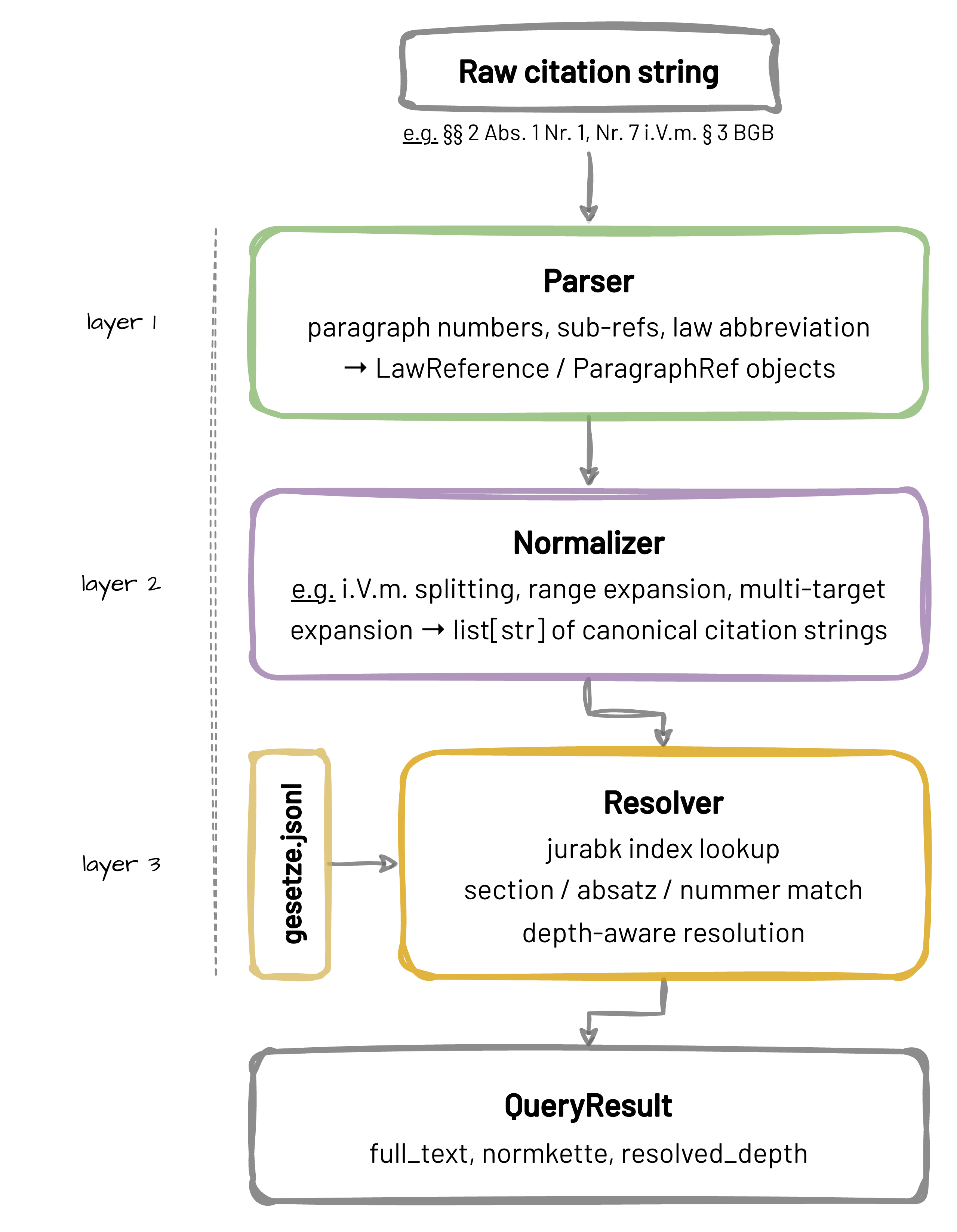}
    \caption{High-level architecture. The parser identifies the structure of a raw citation string, the normalizer expands it into canonical references, and the resolver links these references to statutory provisions in the corpus.}
    \label{fig:architecture}
\end{figure}

Figure~\ref{fig:architecture} 
shows the simplified architecture of \bundesrecht{}. The parser is the first layer. It takes a raw citation string and returns structured reference objects. The normalizer is built on this representation and returns canonical citation strings. 

The resolver then uses these canonical references along with the structured statutory corpus to return the corresponding statutory text.
The resolver is corpus-backed. It loads the JSONL corpus and builds an index over law abbreviations such as \textit{BGB}, \textit{UrhG}, or \textit{ZPO}. During lookup, the resolver first selects the relevant law and then follows the lower-level structure of the reference. The result is returned as a \texttt{QueryResult}, which contains the parsed reference, the resolved law data, the resolved depth, and methods for accessing the resolved text.

\subsection{Parser}
The parser maps a raw citation string to a structured object representation. The main object is \texttt{LawReference}, which contains the law abbreviation and one or more \texttt{ParagraphRef} objects. Each \texttt{ParagraphRef} stores the referenced \textit{Paragraph} or \textit{Artikel} and a list of lower-level references. These lower-level references are then represented as \texttt{SubReference} objects, for example \textit{Absatz}, \textit{Satz}, \textit{Nummer}, \textit{Buchstabe}, \textit{Alternative}, or \textit{Halbsatz}.

Parsing \legalref{\S~81 Abs.\ 1 Nr.\ 1 Buchst.\ a BGB}, e.g., produces the following nested representation:

\begin{lstlisting}[style=examplebox]
LawReference(
  law="BGB",
  is_art=False,
  paragraphs=[
    ParagraphRef(
      paragraph="81",
      sub_refs=[
        SubReference(level="Abs", number="1"),
        SubReference(level="Nr", number="1"),
        SubReference(level="Buchst", number="a")
      ]
    )
  ]
)
\end{lstlisting}

This representation separates the law abbreviation, the referenced \textit{Paragraph}, and the lower-level sub-references. The same object representation also supports references with several \textit{Paragraph} entries, ranges, continuation markers such as \textit{f.}\ and \textit{ff.}, and \textit{Artikel}-based references. It is then used by the normalizer and resolver, but can also be used separately when only reference parsing is needed.

\subsection{Normalizer}
The normalizer builds on the parser output. While the parser identifies the internal structure of a reference, the normalizer uses this structure to produce canonical references that can be resolved independently. It handles several common forms of German statutory references, including \textit{i.V.m.}\ constructions, \textit{\S\S} ranges, \textit{f.}\ and \textit{ff.} references, abbreviated \textit{S.}\ forms for \textit{Satz}, no-space variants such as \legalref{\S312BGB}, and compact multi-target references.

Table~\ref{tab:normalization_examples} shows representative examples. The output is always a list of canonical references, each of which can be passed to the resolver separately.

\begin{table}[h!]
\centering
\small
\renewcommand{\arraystretch}{1.25}
\begin{tabular}{p{0.45\columnwidth}p{0.45\columnwidth}}
\hline
Input & Normalized output \\
\hline
\legalref{\S~312 i.V.m.\ \S~355 BGB} &
\begin{tabular}[t]{@{}l@{}}
\legalref{\S~312 BGB}\\
\legalref{\S~355 BGB}
\end{tabular} \\
\hline

\legalref{\S\S~12--14 BGB} &
\begin{tabular}[t]{@{}l@{}}
\legalref{\S~12 BGB}\\
\legalref{\S~13 BGB}\\
\legalref{\S~14 BGB}
\end{tabular} \\
\hline

\legalref{\S~2 Abs.\ 1 Nr.\ 1, Nr.\ 7, Abs.\ 2 UrhG} &
\begin{tabular}[t]{@{}l@{}}
\legalref{\S~2 Abs.\ 1 Nr.\ 1 UrhG}\\
\legalref{\S~2 Abs.\ 1 Nr.\ 7 UrhG}\\
\legalref{\S~2 Abs.\ 2 UrhG}
\end{tabular} \\
\hline

\legalref{\S~312 f. BGB} &
\begin{tabular}[t]{@{}l@{}}
\legalref{\S~312 BGB}\\
\legalref{\S~313 BGB}
\end{tabular} \\
\hline
\end{tabular}
\caption{Examples of reference normalization in \bundesrecht{}.}
\label{tab:normalization_examples}
\end{table}

The normalizer treats \textit{f.}\ and \textit{ff.}\ differently. A reference with \textit{f.}\ is expanded to the referenced \textit{Paragraph} and the following one. A reference with \textit{ff.}\ is kept unchanged by default, because the number of following \textit{Paragraphen} is not fixed. If an expansion value is provided, \textit{ff.}\ can also be expanded into a fixed number of following \textit{Paragraphen}.

\subsection{Resolver}
The resolver builds on the canonical references produced by the normalizer. It is the corpus-backed part of the library and is exposed through the \texttt{\bundesrechtclass} class, which loads the JSONL corpus once and then resolves references through \texttt{query()} or \texttt{query\_canonical()}. The \texttt{query()} method runs the full workflow: it takes a raw citation string, normalizes it into canonical references, and resolves each reference against the corpus. The \texttt{query\_canonical()} method is used when the input is already canonical and can therefore be resolved directly.

Resolution starts with the law abbreviation. The resolver uses the \texttt{jurabk} field and related metadata to identify the relevant law. It then matches the \textit{Paragraph} or \textit{Artikel} and follows the lower-level structure when the reference contains \textit{Absatz}, \textit{Satz}, \textit{Nummer}, \textit{Buchstabe}, or \textit{Unterbuchstabe}. The returned \texttt{QueryResult} contains the resolved depth, such as section, \textit{Absatz}, \textit{Satz}, \textit{Nummer}, or \textit{Buchstabe}, together with the most specific text found. Appendix~\ref{app:resolver_output} shows an example.

This depth-aware resolution is important because statutory references often point below the level of a \textit{Paragraph} or \textit{Artikel}. Returning the whole \textit{Paragraph} can add text that was not actually cited, so \legalref{\S~312 Abs.\ 2 Nr.\ 7 BGB} should resolve to \textit{Nummer}~7 inside \textit{Absatz}~2, rather than only to \legalref{\S~312 BGB}. If the exact lower-level target cannot be found, the resolver returns the closest available provision together with the resolved level, so downstream systems can distinguish exact from partial matches. For example, a reference to a non-existent \textit{Satz} resolves to the containing \textit{Absatz}, with a note recording the fallback.

\bundesrecht{} is rule-based, so reference processing is fast and requires no model inference. On an Apple M1 laptop, parsing and normalizing a reference takes on average $15.2$ microseconds, measured over all $2{,}944$ evaluation references, while end-to-end resolution through the corpus takes on average $32.8$ microseconds per reference. Resolution additionally requires loading the structured corpus once, which takes about three seconds on first use. These numbers show that the library can process a large number of references efficiently, for example when annotating a full corpus of court decisions.

\section{The \bundesrechtdataset{} Dataset}
The resolver depends on a machine-readable version of German federal law. For this reason, \bundesrecht{} is introduced together with a structured JSONL corpus of the laws. The dataset is derived from \textit{Gesetze im Internet}, the official publication platform for German federal statutory texts, and contains $6{,}873$ German federal laws and regulations in the version used for this paper.

\subsection{Parsing the Source Data}
The corpus is built from the XML files published through \textit{Gesetze im Internet}. For each law, our extraction tool parses the XML file and collects the official abbreviation, title information, publication metadata, source URLs, footnotes, and the list of provisions.

Each provision is stored as a section object. Our tool extracts the provision identifier, for example \legalref{\S~81} or \legalref{Art.~1}, the provision title, and the content below it. Within the content, \textit{Absatz} text, numbered list items, lettered items, \textit{Unterbuchstaben}, and \textit{Listenende} are stored separately when they are present in the original data. This keeps the parts that the resolver needs separate instead of reducing the provision to a single text field.

The resulting dataset is stored as \texttt{gesetze.jsonl}, with one law per line. Each record is identified by a \texttt{gesetze\_id}, which combines the law abbreviation and the XML basename, for example \texttt{BGB::BJNR001950896}.

\subsection{Schema and Hierarchy}
The dataset follows the structure of German statutory drafting. As mentioned above, each law contains metadata, law-level footnotes, source URLs, and a list of sections. Each section stores its identifier, title, content blocks, section-level footnotes, and \textit{Gliederung}. The \textit{Gliederung} preserves higher-level structure such as \textit{Buch}, \textit{Abschnitt}, \textit{Titel}, and \textit{Untertitel} when available.

Inside each section, content is represented through structured blocks containing \textit{Absatz} and, where present, nested \textit{Nummer}, \textit{Buchstabe}, \textit{Unterbuchstabe}, and \textit{Listenende} fields, so a provision is stored as a structured object rather than flat text.

A simplified schema of one JSON object in \texttt{gesetze.jsonl}, representing one law, is shown in Appendix~\ref{app:dataset_schema}. A full entry contains additional metadata and complete section content, but the schema shows the main structure used by the resolver.

\subsection{Corpus Statistics}
The version of the corpus used in this paper contains $6{,}873$ German federal laws and regulations, $107{,}788$ section objects, and $592{,}438$ structured lower-level fields. It also has $154{,}584$ fields in which at least one statutory reference was detected, spread across $5{,}705$ laws and regulations. These numbers show that statutory references are not isolated examples, but occur across a large part of the federal statutory corpus. Detailed corpus statistics are reported in Appendix~\ref{app:corpus_statistics}.

\subsection{Why Structure Matters}
The corpus structure matters because statutory references often point below the level of entire sections. A flat text version can identify that a reference points to \texttt{BGB} or \legalref{\S~312 BGB}, but cannot reliably return only \textit{Absatz}~2 or \textit{Nummer}~7 unless these levels are stored separately, which is the basis for the depth-aware resolution described above.

This structure also makes the corpus easier to use in legal NLP pipelines. A retrieval system can index complete sections, individual \textit{Absätze}, or lower-level fields separately. A question answering system can return the cited part instead of a longer provision. Citation extraction and linking pipelines can also use these identifiers and hierarchy to link references directly to the corresponding statutory provisions.

\section{Evaluation}
We evaluate the parser and normalizer against the German legal reference annotation dataset introduced by \citet{darji2023dataset}. The dataset contains $2{,}944$ manually annotated German legal references. We use this dataset because it provides structured annotation fields for the main parts of German statutory references, including \textit{Artikel}, \textit{Absatz}, \textit{Satz}, \textit{Nummer}, and \textit{Buchstabe}.

The strict and micro evaluations below focus on parsing and normalization. For each raw citation string, we run the normalizer, which internally uses the parser to identify the structure of the reference and then produces canonical references. The extracted field values from these canonical references are compared against the annotated fields. We evaluate the resolver separately in Section~\ref{sec:resolution_evaluation}, since resolution links each reference to statutory text rather than to annotated fields, and no gold dataset links German citations to their resolved text for automatic scoring.

\subsection{Evaluation Setup}
We report two evaluation metrics. The first is a strict exact match at the row level. For each field, the extracted set of values must be identical to the annotated set. A row is considered incorrect if a value is missing or if an extra value is produced.

The second metric uses micro information extraction metrics. For each field, true positives, false positives, and false negatives are accumulated over individual extracted values. This is useful for references with multiple values, for example \legalref{Abs.~1 und 2} or \legalref{Nr.~1, 2, 3}. Rows where both the annotation and the extraction are empty are ignored in this metric, since true negatives are not defined in standard information extraction evaluation.

For the confidence intervals, we use a citation-level bootstrap over the $2{,}944$ evaluation references with $10{,}000$ resamples and report the $2.5$th and $97.5$th percentiles. For system comparisons, we use paired randomization tests with $10{,}000$ permutations on the same references and apply Holm correction across the comparable system-field pairs. All resampling procedures use random seed~42.

\subsection{Strict Exact-Match Evaluation}
Table~\ref{tab:strict_exact_match} shows the strict exact-match results. The match rate is at least $99.7\%$ for all evaluated fields, with \textit{Satz} the highest.

\begin{table}[h!]
\centering
\small
\begin{tabular}{lrrr}
\hline
Field & Correct rows & Match rate & Support \\
\hline
\textit{Artikel} & 2,934 / 2,944 & $99.7\%$ & 2,947 \\
\textit{Absatz} & 2,936 / 2,944 & $99.7\%$ & 2,148 \\
\textit{Satz} & 2,938 / 2,944 & $99.8\%$ & 889 \\
\textit{Nummer} & 2,934 / 2,944 & $99.7\%$ & 478 \\
\textit{Buchstabe} & 2,935 / 2,944 & $99.7\%$ & 29 \\
\hline
\end{tabular}
\caption{Strict exact-match results. Support is the number of annotated values for the field. Correct rows count all rows, including rows where both the annotation and extraction are empty.}
\label{tab:strict_exact_match}
\end{table}

The strict match rate for \textit{Buchstabe}, interpreted together with its support of only 29 annotated values, shows that this field is rare in the dataset: most rows contain no \textit{Buchstabe}, and empty–empty cases count as correct under this strict metric.

\subsection{Micro Information Extraction Metrics}
Table~\ref{tab:micro_ie_metrics} shows the micro information extraction results. The parser and normalizer achieve F1 scores of at least $98.4\%$ for \textit{Artikel}, \textit{Absatz}, \textit{Satz}, and \textit{Nummer}. For the rare \textit{Buchstabe} field, this is more informative because empty rows no longer dominate the result. We also report $95\%$ citation-level bootstrap confidence intervals for micro F1: $99.64$--$99.92$ for \textit{Artikel}, $99.49$--$99.91$ for \textit{Absatz}, $99.24$--$99.89$ for \textit{Satz}, $97.26$--$99.38$ for \textit{Nummer}, and $70.97$--$92.86$ for \textit{Buchstabe}.

\begin{table}[h!]
\centering
\small
\begin{tabular}{lrrrr}
\hline
Field & Precision & Recall & F1 & Support \\
\hline
\textit{Artikel} & $99.9\%$ & $99.7\%$ & $99.8\%$ & 2,947 \\
\textit{Absatz} & $99.9\%$ & $99.6\%$ & $99.7\%$ & 2,148 \\
\textit{Satz} & $100.0\%$ & $99.2\%$ & $99.6\%$ & 889 \\
\textit{Nummer} & $98.9\%$ & $97.9\%$ & $98.4\%$ & 478 \\
\textit{Buchstabe} & $88.5\%$ & $79.3\%$ & $83.6\%$ & 29 \\
\hline
\end{tabular}
\caption{Micro information extraction metrics over individual field values.}
\label{tab:micro_ie_metrics}
\end{table}

To separate the contribution of normalization from parsing, we also compare the direct parser output with the parser followed by normalization on the same $2{,}944$ references. Micro F1 changes from $99.83$ to $99.80$ for \textit{Artikel}, $98.80$ to $99.72$ for \textit{Absatz}, $91.45$ to $99.60$ for \textit{Satz}, $94.43$ to $98.42$ for \textit{Nummer}, and $79.25$ to $83.64$ for \textit{Buchstabe}. The largest improvement is for \textit{Satz}, where normalization expands abbreviated forms such as \textit{S.}; among the $104$ references containing this form, $94$ previously incorrect field sets match the annotation after normalization.

A manual inspection of the mismatched rows shows that disagreements are few and spread across fields rather than concentrated in any single pattern. The mismatches affect only $30$ of the $2{,}944$ rows, accounting for $43$ field-level disagreements, and most are false negatives, where the parser does not produce an annotated value, rather than false positives. The affected rows tend to involve reference forms outside the main patterns of German federal statutory text, such as non-standard notation like \legalref{N1.1} in \legalref{\S~2a Abs.~1 N1.1 ArbGG}, or EU regulation references using abbreviated citation formats. The parser therefore handles the common reference forms reliably, with the remaining mismatches concentrated in atypical cases.

We also test sensitivity to formatting variations using $19{,}826$ spacing, line-break, and marker variants of the evaluation references. No previously correct reference becomes incorrect under field-set scoring after normalization. For references that resolve before perturbation, the returned text also remains unchanged under every applicable transformation. Appendix~\ref{app:formatting_variation} reports the detailed results.

We also examine performance by citation complexity and statute. Among the $234$ references with multiple annotated values, field-level F1 ranges from $92.31\%$ to $99.36\%$ (\textit{Buchstabe} has only seven gold values in this subset). For the two largest statute groups, pooled micro F1 over the five evaluated fields is $99.53\%$ for \textit{ZPO} ($233$ citations) and $99.46\%$ for \textit{BGB} ($225$ citations).

\subsection{Comparison with Prior Work}
\label{sec:tool_comparison}

The evaluation above measures \bundesrecht{} in absolute terms. To place these results in context, we compare against the three most closely related systems introduced in Section~\ref{sec:related}: \texttt{german-legal-reference-parser} (LAVIS), \texttt{legal-reference-extraction} (refex, $0.5.3$), and \textit{Gesetzessuche} ($0.2.0$). None of these produces canonical references, so a direct end-to-end comparison is not possible. We therefore compare at the sub-task shared by all four systems, namely parsing and field extraction. As described in Section~\ref{sec:reference_processing}, reference detection in full text precedes this task.

We run all systems on the same $2{,}944$ annotated references and score their extracted fields with the same information extraction metrics used above. For each system, we add an output adapter that maps its output onto the annotated fields as closely as possible, for example by converting Roman numerals. Rows where a system produces no output are considered false negatives. Adapter details are reported in Appendix~\ref{app:tool_comparison}.

\begin{table}[h!]
\centering
\small
\begin{tabular}{lrrrr}
\hline
Field & Ours & LAVIS & refex & Ges. \\
\hline
\textit{Artikel} & $\mathbf{99.8}$ & $87.5$ & $96.1$ & $98.8$ \\
\textit{Absatz} & $\mathbf{99.7}$ & $82.9$ & -- & $94.1$ \\
\textit{Satz} & $\mathbf{99.6}$ & $85.0$ & -- & $86.8$ \\
\textit{Nummer} & $\mathbf{98.4}$ & $74.4$ & -- & $70.0$ \\
\textit{Buchstabe} & $\mathbf{83.6}$ & -- & -- & $12.9$ \\
\hline
\end{tabular}
\caption{Micro F1 per field over all rows, in percent. A dash means the system does not produce the field: refex outputs only the provision number, and LAVIS has no \textit{Buchstabe} field. \textit{Ges.}\ is \textit{Gesetzessuche}.}
\label{tab:tool_comparison}
\end{table}

Table~\ref{tab:tool_comparison} shows that \bundesrecht{} obtains the highest micro F1 on every comparable field. Paired randomization tests with Holm adjustment give $p \leq 0.002$ for all comparable system-field pairs in the all-row evaluation. The systems also differ in coverage, which we define as producing at least one value in the fields compared for a given system. \bundesrecht{} produces output for $99.9\%$ of the references, LAVIS for $78.0\%$, refex for $92.7\%$, and \textit{Gesetzessuche} for $100\%$. When \bundesrecht{} and a competing system are scored only on the rows for which that system produces an output, and averaged across that system's fields, \bundesrecht{} reaches a macro F1 of $99.6$ against $96.2$ on the four LAVIS fields, $99.9$ against $99.8$ on the single refex field, and $96.2$ against $72.5$ on the five \textit{Gesetzessuche} fields. On these coverage-adjusted rows, the \textit{Artikel} differences against LAVIS and refex are not significant after Holm adjustment ($p=1.0$ for both).

A few factors explain most of these gaps. LAVIS produces no output for $22.0\%$ of the annotated rows. \textit{Gesetzessuche} returns only one parsed reference per call and therefore misses the additional targets in multi-reference expressions such as \legalref{\S~2 Abs.~1 Nr.~1, Nr.~7, Abs.~2 UrhG}. No other tool performs strongly across all five fields, since refex outputs only the provision number, LAVIS has no \textit{Buchstabe} field, and \textit{Gesetzessuche} reaches only $12.9$ F1 on it. In contrast, \bundesrecht{} performs strongly across all fields under both comparisons, and, as before, only the rare \textit{Buchstabe} field remains harder to evaluate.

\subsection{Resolution Evaluation}
\label{sec:resolution_evaluation}
The evaluations above measure parsing and normalization. We now evaluate the third step, resolution, which links a canonical reference to the corresponding statutory text in the corpus. No gold dataset links German statutory citations to their resolved text; therefore, we manually verify the resolver's output against the official statutory text. 

We draw a stratified sample of $200$ citation strings across the six reference levels from the observed surface forms in the Open Legal Data corpus~\citep{darji2025segmentation}, the same pool used in Section~\ref{sec:dedup}. Many of these strings are multi-target, so after normalization the sample expands to $433$ individually resolvable targets. For each target, we locate the referenced provision on \textit{Gesetze im Internet} and check whether the text returned by the resolver matches the unit shown there. Citations whose target has since been repealed or renumbered, and so is absent from the current corpus, are excluded, since these reflect the historical coverage of the snapshot rather than a resolution error. 

\begin{table}[h!]
\centering
\small
\begin{tabular}{lrr}
\hline
Level & Correct / Total & Accuracy \\
\hline
\textit{Section} & \hphantom{0}98 / \hphantom{0}98 & $100.0\%$ \\
\textit{Absatz} & 104 / 104 & $100.0\%$ \\
\textit{Nummer} & \hphantom{0}36 / \hphantom{0}38 & $94.7\%$ \\
\textit{Satz} & \hphantom{0}75 / \hphantom{0}90 & $83.3\%$ \\
\textit{Buchstabe} & \hphantom{0}94 / 101 & $93.1\%$ \\
\textit{Unterbuchstabe} & \hphantom{00}2 / \hphantom{00}2 & $100.0\%$ \\
\hline
\end{tabular}
\caption{Per-level resolution accuracy over the $433$ verified targets.}
\label{tab:resolution}
\end{table}

Table~\ref{tab:resolution} reports per-level resolution accuracy. Overall, $409$ of the $433$ targets resolve correctly ($94.5\%$). Resolution is exact at the \textit{Section}, \textit{Absatz}, and \textit{Unterbuchstabe} levels and near-exact at the \textit{Nummer} and \textit{Buchstabe} levels. \textit{Satz} is the most difficult level, at $83.3\%$.

Of the $24$ resolution errors, $22$ occur in the resolver. Fifteen of these are sentence-retrieval failures caused by embedded lists, where sentence counting cannot descend into the list or resume correctly afterward, and the remaining seven resolver errors occur when a compact citation omits an intervening \textit{Absatz} or \textit{Nummer}, causing resolution to fall back to a broader provision. One error comes from parsing, where \textit{Nr.~4a} is interpreted as \textit{Nummer~4} with \textit{Buchstabe~a}, and another from corpus structuring, where a flattened list prevents lookup of the requested item.

\section{Canonical Reference Deduplication}
\label{sec:dedup}

The evaluation earlier measures whether the parser and normalizer extract the correct reference fields. However, many downstream legal NLP tasks often need one more step: they need to know when different citation strings refer to the same statutory target. Na\"ive string matching is insufficient for this task, because the same target may use different spacing, punctuation, or abbreviations.

We therefore evaluate whether canonical references can be used to group different surface forms of the same statutory reference. Starting from the distinct raw citation surface forms in the \texttt{references.law} field of the Open Legal Data corpus~\citep{darji2025segmentation}, a rule-based labeling procedure assigns high-confidence single-provision references to canonical targets, and we keep the targets with at least two observed surface forms. A successful normalizer should map all surface forms in such a group to the same canonical target. Over these groups, \bundesrecht{} matches the silver-labeled canonical target for $98.99\%$ of surface forms. Appendix~\ref{app:canonical_reference_deduplication} reports the construction details and counts.

We also evaluate deduplication directly. Two surface forms should be grouped together if and only if they share the same canonical target. We compare a surface-string baseline, which keys on citation strings after only lowercasing and whitespace normalization, against canonical-reference matching using the output of \bundesrecht{}. The results in Table~\ref{tab:canonical_reference_deduplication} show that surface-string matching yields very low recall because surface variants remain separated. Canonical-reference matching provides much higher recall while maintaining precision, showing that normalization makes citation strings easier to group, compare, and deduplicate. Since the surface-string baseline applies no reference normalization, this comparison also acts as an ablation of the normalization step.

\begin{table}[h!]
\centering
\small
\begin{tabular}{lrrr}
\hline
Matching key & Precision & Recall & F1 \\
\hline
Raw surface string & $53.0\%$ & $0.2\%$ & $0.4\%$ \\
Canonical reference & $100.0\%$ & $\boldfw{99.0\%}$ & $99.5\%$ \\
\hline
\end{tabular}
\caption{Deduplication results over citation surface variants. Canonical-reference matching precision is \(100.0\%\), with no false-positive pairs on the evaluated silver-standard subset.}
\label{tab:canonical_reference_deduplication}
\end{table}

This experiment also complements the field-level evaluation on a second, independent dataset. The manually annotated reference dataset measures whether fields such as \textit{Artikel}, \textit{Absatz}, \textit{Satz}, \textit{Nummer}, and \textit{Buchstabe} are extracted correctly. The deduplication evaluation instead measures whether these fields are combined into a canonical reference that can group real surface variants of the same statutory target. This is important for corpus construction and citation-based analysis, where the same provision may otherwise be counted as several unrelated strings.

\section{Applications}
\bundesrecht{} turns citation strings into structured corpus annotations. Instead of storing only the surface form found in a court decision or a legal document, a corpus can store the cited law, provision, lower-level unit, resolved depth, and resolved text, making later processing less dependent on how a citation was written.

The same representation is useful for retrieval, legal question answering, retrieval-augmented generation, and citation-based analysis. Depth-aware resolution can return the cited lower-level unit instead of a whole section, reducing unnecessary context, while canonical references make it possible to treat statutory references as links between provisions. This can show which provisions are frequently cited, which laws depend heavily on others, and which provisions should be reviewed when a highly cited law is amended.

\section{Conclusion}
This paper introduces \bundesrecht, an open resource for German statutory reference processing that combines an open-source software library with a structured JSONL corpus of German federal law. The library supports parsing, normalization, and resolution of statutory references, including compact and compound forms such as \legalref{\S~312 i.V.m.\ \S~355 BGB}, ranges, and multi-target references. The corpus maintains the hierarchy of statutes down to lower-level fields such as \textit{Nummer}, \textit{Buchstabe}, and \textit{Unterbuchstabe}, which makes depth-aware resolution possible.

We evaluated the parser and normalizer on $2{,}944$ manually annotated references using strict exact-match and micro information extraction metrics, and compared against relevant prior work, with strong performance across the main reference fields. We additionally verified resolution against the official statutory text. A canonical reference deduplication experiment further shows that normalization maps real citation surface variants to shared canonical targets. The parser and normalizer are modular and may be adaptable to other statutory systems, but we leave such extensions to future work because this would require adjusting citation patterns and reference conventions. Together, the library and corpus provide a reusable resource for German legal NLP systems that need statutory references in a structured, comparable, and resolvable form.

\section*{Limitations}
The main limitation of the current corpus is the \textbf{table structure}. Some statutes contain tables, especially in annexes, fee schedules, and technical regulations. In the current dataset, these are converted into plain text inside content blocks, so row and column boundaries are not preserved. This affects use cases where the exact cell matters, for example when a fee, threshold, or technical value belongs to a specific row and column. The limitation affects a small part of the corpus, as tables occur in $4.6\%$ of section objects and $1.9\%$ of non-empty content blocks.

Another limitation is \textbf{historical coverage}. The published corpus is a snapshot of the XML files available from \textit{Gesetze im Internet} at extraction time and does not contain historical versions of provisions. As a result, the resolver links references to the version present in the corpus, not necessarily the version in effect when a citation was originally written. This matters for legal-historical analysis and for older documents.

The parser and normalizer work on citation surface forms and could potentially be adapted to other statutory systems, such as Austrian or Swiss federal law or EU legislation, where some reference conventions overlap. Such an extension would still require adjusting citation patterns, abbreviations, and jurisdiction-specific reference rules. Resolution would also require a structured statutory corpus and mappings to the hierarchy used in the target legal system. We have not evaluated transfer outside German federal law.

\section*{Ethical Considerations}
Our resource is based on German federal statutory text published through \textit{Gesetze im Internet}. The dataset does not contain personal data or court-party information, and consists of publicly available legal text and source metadata derived from the official XML files. The \bundesrecht{} library is released under the MIT License, and our corpus data under the Open Database License (ODbL).

During data collection, the download script identifies itself through a custom user agent containing the project name, institutional context, research purpose, and contact information. We also rate-limit downloads with a randomized delay between requests, to make the collection process transparent and avoid unnecessary load on the source website.

\bibliography{custom}

\appendix

\section{Resolver Output Example}
\label{app:resolver_output}
For the reference \legalref{\S~312 Abs.\ 2 Nr.\ 7 BGB}, the resolver returns a \texttt{QueryResult} whose resolved depth is \textit{Nummer}:

\begin{lstlisting}[style=examplebox]
QueryResult(
  reference="§ 312 Abs. 2 Nr. 7 BGB",
  resolved_para="312 Abs. 2 Nr. 7",
  resolved_depth="nummer",
  titel="Anwendungsbereich",
  text="Behandlungsverträge nach § 630a,"
)
\end{lstlisting}

\section{Simplified Dataset Schema}
\label{app:dataset_schema}

Figure~\ref{fig:dataset_schema} shows the simplified JSON structure of one law record in \texttt{gesetze.jsonl}.

\begin{figure}[h!]
\centering
\begin{lstlisting}[style=examplebox]
{
  "gesetze_id": string,
  "jurabk": string,
  "metadaten": { ... },
  "fussnoten": [string],
  "quelle": { ... },
  "sections": [
    {
      "paragraf": string,
      "titel": string,
      "content": [
        {
          "absatz": string,
          "nummer": [
            {
              "text": string,
              "buchstaben": [
                {
                  "text": string,
                  "unterbuchstaben": [
                    {"text": string}
                  ],
                  "listenende": [{...}]
                }
              ],
              "listenende": [{...}]
            }
          ],
          "listenende": [
            {
              "level": string,
              "start": integer,
              "end": integer or null,
              "text": string
            }
          ]
        }
      ],
      "fussnoten": [string],
      "gliederung": [
        {
          "gliederungsbez": string,
          "gliederungstitel": string
        }
      ]
    }
  ]
}
\end{lstlisting}
\caption{Simplified schema of one JSON object in \texttt{gesetze.jsonl}.}
\label{fig:dataset_schema}
\end{figure}

\section{Detailed Corpus Statistics}
\label{app:corpus_statistics}

Table~\ref{tab:corpus_statistics_appendix} summarizes the corpus size for the version used in this paper. We report the number of laws, section objects, structured fields, reference-bearing fields, \textit{Gliederung} entries, and footnotes to show both the scale of the corpus and the amount of structure preserved during extraction.

\begin{table}[h!]
\centering
\small
\begin{tabular}{lr}
\hline
Statistic & Count \\
\hline
Laws and regulations & 6,873 \\
Section objects & 107,788 \\
Structured fields & 592,438 \\
Fields containing statutory references & 154,584 \\
Laws with reference-bearing fields & 5,705 \\
\textit{Gliederung} entries & 125,978 \\
Footnotes & 26,407 \\
\hline
\end{tabular}
\caption{Detailed statistics of the structured \bundesrechtdataset{} corpus.}
\label{tab:corpus_statistics_appendix}
\end{table}

The structured fields consist of $280{,}223$ \textit{Absatz} fields, $231{,}261$ \textit{Nummer} fields, $53{,}126$ \textit{Buchstabe} fields, $5{,}448$ \textit{Unterbuchstabe} fields, and $22{,}380$ \textit{Listenende} fields. Fields containing statutory references are structured fields in which at least one statutory reference was detected. Laws with reference-bearing fields count how many laws contain at least one such field.

\section{System Comparison Details}
\label{app:tool_comparison}

This appendix describes how each system in Section~\ref{sec:tool_comparison} is run and mapped onto the annotated fields. All systems are evaluated on the same $2{,}944$ annotated references and scored with the same code as \bundesrecht{}. Each is pinned to a fixed version: LAVIS at commit \texttt{66f875f}, refex at $0.5.3$, and \textit{Gesetzessuche} at $0.2.0$. The published LAVIS package omits a required resource file from its wheel, which we restored (\texttt{resource/laws.txt}) from the pinned commit without modifying any parser logic. 

\paragraph{Field mapping.}
Each system exposes its own output format, so we added a thin adapter that maps this output onto the annotated fields \textit{Artikel}, \textit{Absatz}, \textit{Satz}, \textit{Nummer}, and \textit{Buchstabe}. The adapters normalize values only in ways that do not change their identity: they strip and collapse whitespace and lowercase the result. Roman sub-reference values from \textit{I} to \textit{XII} are mapped to their Arabic form, since the two are equivalent in the annotated data. Adapters do not add, split, or infer values beyond what the tool produces, except where stated below.

\paragraph{LAVIS.}
LAVIS parses a reference into structured objects and exposes each component of multi-target and \textit{i.V.m.}\ references. We map its \textit{Paragraph} to \textit{Artikel}, and its \textit{Absatz} and \textit{Satz} to the matching fields. LAVIS reports \textit{Nummer} and lettered sub-references through a single field, so when a value combines a number and a letter, such as \legalref{1a}, only the \textit{Nummer} part is scored, since LAVIS has no separate \textit{Buchstabe} field. Purely alphabetic values cannot be assigned to a native field and are not scored. LAVIS is therefore compared on \textit{Artikel}, \textit{Absatz}, \textit{Satz}, and \textit{Nummer}.

\paragraph{refex.}
refex extracts references from text and reports the cited provision number, which corresponds to the \textit{Paragraph} or \textit{Artikel}, so we map it to \textit{Artikel}. refex does not report lower-level fields and is compared on \textit{Artikel} only.

\paragraph{Gesetzessuche.}
\textit{Gesetzessuche} parses a single reference and exposes \texttt{paragraph}, \texttt{section}, \texttt{sentence}, \texttt{number}, and \texttt{letter} fields, which we map to \textit{Artikel}, \textit{Absatz}, \textit{Satz}, \textit{Nummer}, and \textit{Buchstabe}, respectively, so it is compared on all five fields. Because it parses only one reference per call, it returns a single target even when the citation string contains several, which lowers its results on multi-target references.

\paragraph{Coverage.}
A reference is counted as covered by a system if that system produces at least one value in the fields compared for that system. Rows for which a given system produces no output are scored as false negatives in the all-rows comparison in Table~\ref{tab:tool_comparison}, and are excluded in the coverage-adjusted comparison, where each system is scored only on the rows it covers.

\section{Canonical Reference Deduplication Statistics}
\label{app:canonical_reference_deduplication}

The labeling procedure used to build the evaluation data is separate from \bundesrecht{} and applies a conservative rule set that retains only simple, high-confidence single-provision references, so the labels serve as a reliable silver standard. Starting from $550{,}657$ distinct raw citation surface forms, it retains $248{,}716$ surface forms and assigns them to $199{,}249$ canonical targets. The deduplication subset contains $35{,}010$ target groups covering $84{,}477$ surface forms, of which \bundesrecht{} matches $83{,}625$. Table~\ref{tab:canonical_reference_data} summarizes these counts.

\begin{table}[h!]
\centering
\small
\begin{tabular}{lr}
\hline
Item & Count \\
\hline
Distinct raw citation surface forms & 550,657 \\
Silver-labeled surface forms & 248,716 \\
Canonical targets & 199,249 \\
Target groups & 35,010 \\
Surface forms in target groups & 84,477 \\
Matched canonical targets & 83,625 \\
\hline
Target agreement & $98.99\%$ \\
\hline
\end{tabular}
\caption{Canonical reference evaluation. Target groups are canonical targets with at least two observed surface forms. The largest target group contains 16 surface forms.}
\label{tab:canonical_reference_data}
\end{table}

The largest target group maps to \legalref{\S~113 Abs.~1 Satz~1 VwGO} and contains $16$ observed surface forms. Other large groups include \legalref{\S~540 Abs.~1 Satz~1 Nr.~1 ZPO} with $15$ surface forms, \legalref{\S~86b Abs.~1 Satz~1 Nr.~2 SGG} with $14$ surface forms, and \legalref{\S~7 Abs.~1 Satz~2 Nr.~2 SGB~2} with $13$ surface forms. These examples show that the variation comes from real citation strings, including spacing, punctuation, abbreviated forms such as \textit{S.} and \textit{Satz}, and law-name variants.

\section{Formatting Variation Analysis}
\label{app:formatting_variation}

We generated $19{,}826$ spacing, line-break, and marker variants of the $2{,}944$ evaluation references using the existing pipeline. At the parsing and normalization stage, no reference that was previously correct became incorrect under field-set scoring. Canonical strings changed for $27$ variants: $12$ whitespace variants, $9$ marker variants, and $6$ combined variants.

All $27$ affected variants were initially unresolved. After perturbation, $25$ remained unresolved, while two marker-expansion variants of the same citation became resolvable. For references that returned text before perturbation, the resolved text remained unchanged under every applicable transformation. These results are a consistency check on formatting variation rather than an estimate of overall coverage or accuracy.

\end{document}